\documentclass[sigconf]{acmart}
\AtBeginDocument{%
  }

\newcommand{\model}{OOD-GraphLLM }
\newcommand{\modelnosp}{OOD-GraphLLM}
\newcommand{\spara}[1]{\smallskip\noindent\textbf{#1}.}

\usepackage{xcolor}
\usepackage{array}
\usepackage{tcolorbox}
\usepackage{multirow}
\usepackage{lipsum}
\usepackage{graphicx}
\usepackage{enumitem}
\usepackage{makecell}
\usepackage{mhchem}
\usepackage{ragged2e}
\usepackage{seqsplit}
\usepackage{tabularx}
\usepackage{pifont}

\newcommand{\smiles}[1]{\begingroup\small\ttfamily\seqsplit{#1}\endgroup}

\newcolumntype{Y}{>{\RaggedRight\arraybackslash}X}

\newcolumntype{C}[1]{>{\centering\arraybackslash}m{#1}}

\begin{document}

\title{OOD-GraphLLM: Graph Large Language Model for Out-of-Distribution Generalized Drug Synergy Prediction}

\author{Xin Wang}
\affiliation{%
  \institution{DCST, BNRist, Tsinghua University}
  \city{Beijing}
  \country{China}}
\email{xin_wang@tsinghua.edu.cn}

\author{Linxin Xiao}
\affiliation{%
  \institution{DCST, Tsinghua University}
  \city{Beijing}
  \country{China}}
\email{xlx21@mails.tsinghua.edu.cn}

\author{Yang Yao}
\affiliation{%
  \institution{DCST, Tsinghua University}
  \city{Beijing}
  \country{China}}
\email{yaoyang21@mails.tsinghua.edu.cn}

\author{Wenwu Zhu}
\authornote{Corresponding author. DCST is the abbreviation for Department of Computer Science and Technology. BNRist is the abbreviation for Beijing National Research Center for Information Science and Technology.} 
\affiliation{%
  \institution{DCST, BNRist, Tsinghua University} 
  \city{Beijing}
  \country{China}}
\email{wwzhu@tsinghua.edu.cn}

\renewcommand{\shortauthors}{Xin Wang et al.}


\begin{abstract}

Drug synergy prediction (DSP) aims to identify efficacious drug combinations under various cellular contexts with different targets. However, the continual emergence of novel compounds results in variations in molecular scaffolds and sizes, causing drug synergy data to exhibit out-of-distribution (O.O.D.) shifts with respect to topological structure. Existing works rely on in-distribution (I.D.) assumption, 
failing to handle the O.O.D. shifts. To solve this problem, we study out-of-distribution generalized drug synergy prediction through a graph large language model for the first time. 
Nevertheless, O.O.D. generalized DSP is highly non-trivial, posing several challenges: i) how to discover structurally relevant and irrelevant molecular representations with respect to cell targets; ii) how to find the optimal graph neural architectures that accurately calculate molecular representations; and iii) how to jointly leverage molecular structural and semantic information in LLMs.
To address these challenges, we propose \textbf{\modelnosp}, a novel graphLLM framework which is able to accurately predict drug synergy under O.O.D. settings via jointly optimizing molecular graph representation and biomedical semantic language representations in a unified manner.
Concretely, we first propose a target-adaptive disentangled molecular graph encoding model to distinguish target-relevant and target-irrelevant molecular representations for both seen and unseen drugs, then introduce a pairwise attentive graph architecture search algorithm that dynamically finds the best neural architectures to calculate molecular representations for different and new drug pairs, followed by our design of multi-level contextualized cellular feature alignment mechanism to incorporate cell line context information at both structural and semantic levels. Furthermore, we finetune DrugSyn-LLM, a biomedical LLM, and employ a retrieval-augmented biomedical instruction tuning strategy to align molecular topological information and molecular semantic information with language-based reasoning for O.O.D. generalized DSP.
Extensive experiments under several O.O.D. settings demonstrate that the proposed \model consistently outperforms state-of-the-art approaches on various DSP tasks.
Both the source code \footnote{\url{https://github.com/EkkoXiao/Bio-GraphLLM}} 
and released model \footnote{\url{https://mn.cs.tsinghua.edu.cn/bio-graphllm/}} are publicly available, where users are allowed to download model resources and interactively use the system through a web interface.

\end{abstract}



\maketitle

\section{Introduction}
Drug combination therapy~\cite{survey-drug-combination-therapies} has emerged as a promising strategy for treating complex diseases such as cancer and drug-resistant infections. Compared to single-drug treatments, effective drug combinations can enhance therapeutic efficacy, reduce toxicity, and mitigate the development of resistance.
Therefore, Drug synergy prediction (DSP) ~\cite{review-drug-synergy-prediction,DeepSynergy,MatchMaker,TreeCombo} has become a critical and fundamental problem in computational drug discovery, aiming to identify efficacious drug combinations under various cellular contexts with different cell targets.
Recent advances in machine learning, particularly graph neural networks (GNNs)~\cite{DeepDDS,DDoS,GAECDS}, have substantially improved DSP by conducting representation learning over the molecular topological structures of drugs to predict drug-drug synergies. 
By representing drugs as molecular graphs, existing literature is able to capture topological structural information that plays an important role in determining the corresponding molecular chemical properties.

\begin{figure}[htbp]
  \centering
  \includegraphics[width=1.0\linewidth]{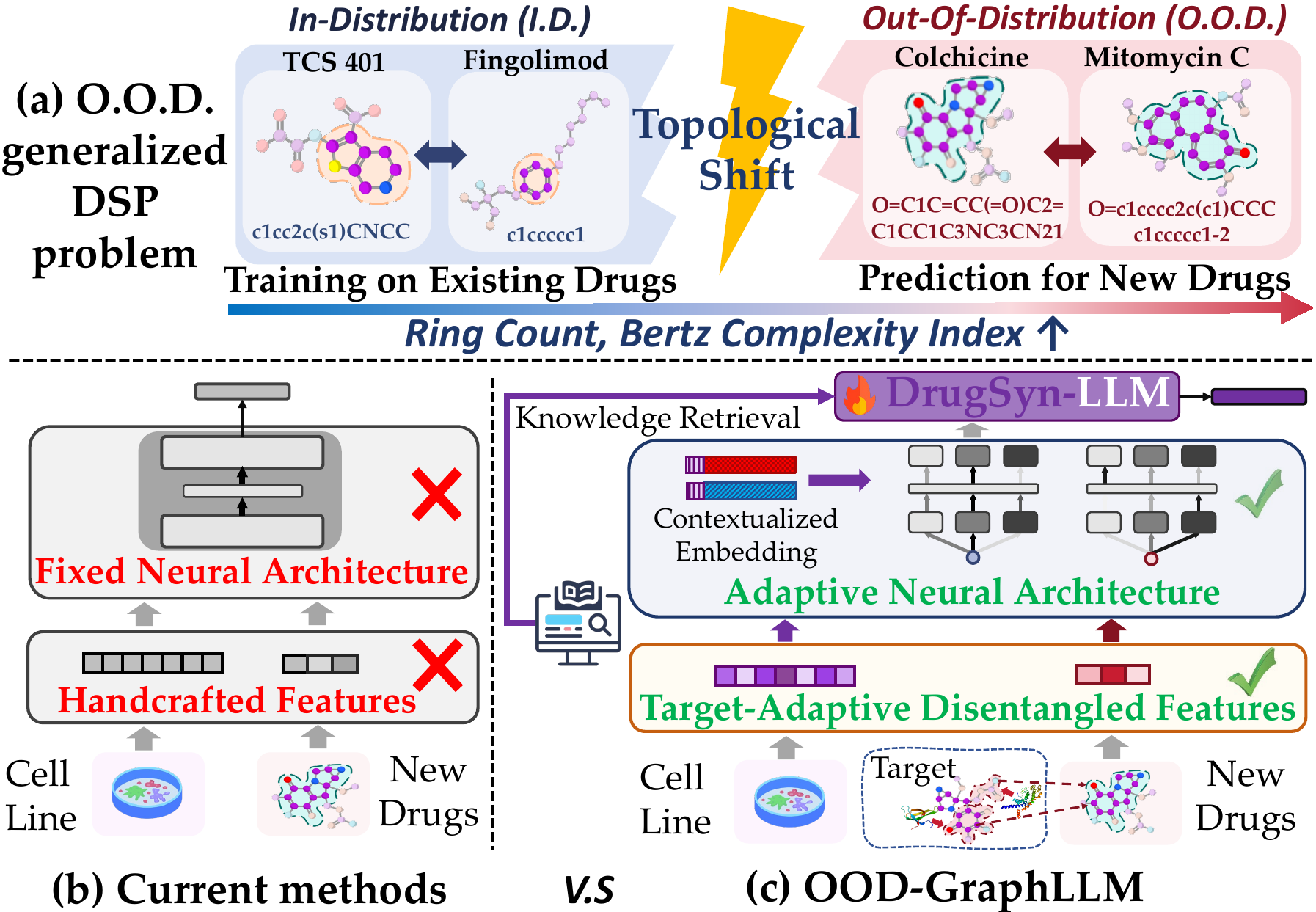}
  \vspace{-5mm}
  \caption{Comparisons between current methods (b) and \model (c) under O.O.D. generalized DSP (a) settings.}
  \label{fig:case}
\end{figure}

However, due to the continuous emergence of novel compounds in drug discovery, drug synergy data often exhibit out-of-distribution (O.O.D.) shifts with respect to topological structure at the drug level, primarily caused by variations in molecular scaffolds and compound sizes. This out-of-distribution generalized drug synergy prediction (O.O.D. generalized DSP) problem requires models to generalize well to new drugs and previously unseen molecular scaffolds. Fig.~\ref{fig:case} (a) illustrates one example of O.O.D. generalized DSP. 
Existing works on DSP heavily rely on the in-distribution (I.D.) assumption, where the drug structures tend to remain the same for training and testing data, failing to handle the O.O.D. shifts.



In this paper, we study out-of-distribution generalized drug synergy prediction by resorting to a graph large language model (shown in Fig.~\ref{fig:case} (c)), to the best of our knowledge, for the first time. Nevertheless, O.O.D. generalized DSP is highly non-trivial, with several key challenges. For drugs with O.O.D. topological structures, 
\begin{enumerate}[leftmargin=0.5cm]
    \item it is challenging to obtain structurally relevant and irrelevant molecular representations with respect to cell targets;
    \item it is challenging to find the optimal graph neural architecture that can calculate accurate molecular representations;
    \item it is challenging to jointly leverage structural and semantic information from molecules within LLMs.
\end{enumerate}



To address these challenges, we propose \textbf{\modelnosp}, a novel graphLLM framework capable of accurately predicting drug synergy under O.O.D. setting via joint optimization of molecular graph representation and biomedical semantic language representations in a unified manner. 
Given new drugs, the proposed \model treats cell lines as contexts, obtains the best graph neural network (GNN) architecture for calculating the new molecular graph representations, aligns these representations with both topological and semantic cellular features, tokenizes all the features as input to a biomedically finetuned LLM for O.O.D. generalized DSP, and optimizes the whole procedure within one single framework.

In concrete, we first propose a \textbf{target-adaptive disentangled molecular graph encoding model} that learns target-relevant and target-irrelevant molecular representations, conditioning the target-relevant representations on various cell targets with disentanglement constraint to preserve target-aware information for new molecular structures.
Building upon the target-adaptive disentangled molecular graph representations, we introduce a \textbf{pairwise attentive graph architecture search algorithm} that dynamically finds the optimal graph neural architectures for new drug pairs, allowing for accurate molecular representation learning under distribution shifts.
We then design a \textbf{multi-level contextualized cellular feature alignment mechanism} to incorporate cell line information into molecular graph representations as contexts at both the structural and semantic levels.
Last but not least, we finetune DrugSyn-LLM, a biomedical LLM, with \textbf{retrieval-augmented biomedical instruction tuning strategy}, aligning molecular topological and semantic information with language-based reasoning to accomplish O.O.D. generalized DSP. We conduct extensive experiments to demonstrate the superiority of \model over state-of-the-art baselines under various O.O.D. settings. The contributions of this paper are summarized as follows:

\begin{itemize}[leftmargin=0.5cm]
    \item We are the first to study the problem of out-of-distribution generalized drug synergy prediction (O.O.D. generalized DSP) by resorting to graph large language models, to the best of our knowledge. 
    \item We propose \textbf{\modelnosp}, a novel graphLLM framework for accurate O.O.D. generalized DSP, which jointly optimizes our proposed four components, i.e., (1) target-adaptive disentangled molecular graph encoding, (2) pairwise attentive graph architecture search, (3) multi-level contextualized cellular feature alignment and (4) finetuned biomedical LLM DrugSyn-LLM with retrieval-augmented instruction tuning, within a unified framework. 
    \item We conduct extensive experiments under multiple O.O.D. evaluation settings to demonstrate that the proposed \model is able to consistently outperform state-of-the-art baselines, highlighting its superior generalization ability and prediction accuracy under molecular topological distribution shifts.
\end{itemize}

\vspace{-6pt}

\section{Related Works}

\spara{Drug Synergy Prediction (DSP)}
Deep learning-based models have been widely adopted for drug synergy prediction due to their ability to model complex drug--drug and drug--cell interactions. 
Early methods, such as DeepSynergy~\cite{DeepSynergy}, MatchMaker~\cite{MatchMaker}, and TreeCombo~\cite{TreeCombo}, mainly used molecular descriptors and cell-line gene expression profiles with deep neural networks or gradient boosting models. 
With the development of graph neural networks (GNNs), methods such as DeepDDS~\cite{DeepDDS}, DDoS~\cite{DDoS}, and GAECDS~\cite{GAECDS} represented drugs as molecular graphs to capture atom-level structural information. 
Meanwhile, DFFNDDS~\cite{DFFNDDS} exploited pretrained language models such as BERT~\cite{Bert} to extract semantic representations from SMILES strings, while DTSyn~\cite{DTSyn} and AttenSyn~\cite{AttenSyn} further introduced attention-based Transformer architectures for drug--drug and drug--cell interaction modeling. 
More recently, BAITSAO~\cite{BAITSAO} has explored large language models as predictors for drug synergy tasks. 
Despite these advances, most existing methods are developed under the in-distribution assumption and are rarely designed for O.O.D. generalization, leaving the integration of GNNs and LLMs for distribution-shift-aware drug synergy prediction underexplored.

\spara{Graph Large Language Models (GraphLLMs)}
Graph large language models (GraphLLMs) extend the reasoning and generation abilities of LLMs to graph-structured data, enabling tasks such as graph understanding and question answering~\cite{GLLMa,GLLMb,GLLMc}. 
Existing studies mainly follow two directions. 
Prompt-based methods, such as InstructGLM~\cite{InstructGLM} and NLGraph~\cite{NLGraph}, translate graph structures into textual prompts that can be interpreted by LLMs. 
Representation-alignment methods, such as GraphGPT~\cite{GraphGPT} and GraphLLM~\cite{GraphLLM}, encode graphs with GNNs and feed the resulting graph tokens into language models. 
Further works, including GLEM~\cite{GLEM} and PATTON~\cite{PATTON}, explore iterative co-training and alignment between GNNs and LLMs to improve representation learning. 
Motivated by the graph structure of molecules and the semantic information in SMILES, recent studies such as MolTC~\cite{MolTC} and DyNAS-DDI~\cite{DyNAS-DDI} have applied GraphLLM-style architectures to drug-related tasks. 
However, existing GraphLLM frameworks are mostly designed for constrained prediction or reasoning scenarios, and have not fully addressed complex drug synergy prediction settings that require modeling higher-order drug--drug--cell context and O.O.D. shifts.

\section{OOD-GraphLLM}

\begin{figure*}
  \centering
  \includegraphics[width=1\textwidth]{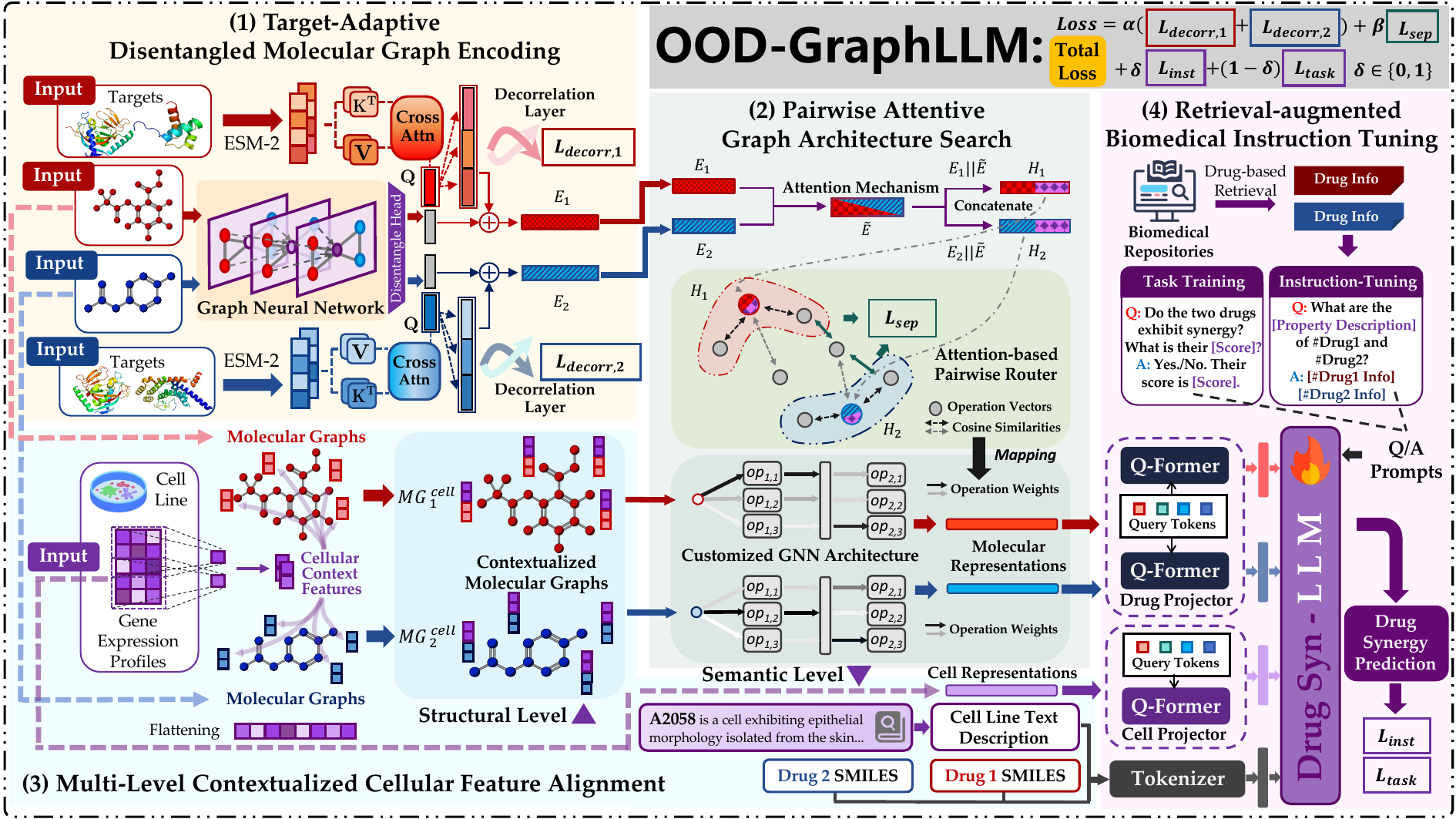}
  \caption{The overall framework of \model. \model is able to conduct accurate O.O.D. generalized DSP by integrating cellular contextualized molecular graph representation learning and LLMs together through four jointly optimized components, i.e., (1) Target-Adaptive Disentangled Molecular Graph Encoding; (2) Pairwise Attentive Graph Architecture Search; (3) Multi-Level Contextualized Cellular Feature Alignment; and (4) Finetuned DrugSyn-LLM with Retrieval-Augmented Biomedical Instruction Tuning.}
  \Description{}
  \label{fig:framework}
\end{figure*}

In this section, we describe the proposed \modelnosp in detail. Sec \ref{module0} formally formulates the drug synergy prediction problem and defines the learning objectives. Sec \ref{module1} and Sec \ref{module2} present the target-adaptive disentangled molecular graph encoding model and the pairwise attentive graph architecture search algorithm, respectively. Sec \ref{module3} describes the multi-level contextualized cellular feature alignment mechanism, and finally, Sec \ref{module4} details the design of finetuning DrugSyn-LLM with the retrieval-augmented biomedical instruction tuning strategy and outlines the overall multi-stage training procedure. Fig.~\ref{fig:framework} shows the overall framework of \modelnosp.

\subsection{Problem Formulation}
\label{module0}

\textbf{Drug synergy prediction (DSP)} aims to characterize the combined effect of multiple drugs under a specific cell line context. Take the most common setting, i.e., DSP for two drugs, as an example, we formally define DSP as follows. Let $\mathcal{D} = \{d_1, d_2, \ldots, d_n\}$ denote the universe of drug molecules and let $\mathcal{C} = \{c_1, c_2, \ldots, c_m\}$ denote the set of cell lines representing distinct biological environments. Each data instance is characterized by a triplet $(d_i, d_j, c_k)$, where $d_i, d_j \in \mathcal{D}$ are two drugs administered in combination and $c_k \in \mathcal{C}$ specifies the cellular condition. The learning objective is to infer a predictive function $f$ as follows:
{
\small
\begin{equation}
f: (d_i, d_j, c_k) \mapsto (y_{ij}^k, s_{ij}^k),
\end{equation}
}
where $y_{ij}^k \in \mathcal{Y}$ denotes a discrete interaction label, and $s_{ij}^k \in \mathbb{R}$ represents a continuous synergy score quantifying the strength of the combinatorial effect. 

\textbf{Out-of-distribution generalized drug synergy prediction (O.O.D. generalized DSP)} studies generalization beyond observed drug distributions. 
Specifically, the drug space $\mathcal{D}$ is partitioned into an in-distribution subset $\mathcal{D}_{\text{I.D.}}$ and an out-of-distribution subset $\mathcal{D}_{\text{O.O.D.}}$, according to criteria such as molecular scaffolds and sizes etc. These subsets satisfy $\mathcal{D}_{\text{I.D.}} \cap \mathcal{D}_{\text{O.O.D.}} = \emptyset$ and $\mathcal{D}_{\text{I.D.}} \cup \mathcal{D}_{\text{O.O.D.}} = \mathcal{D}$.

Under this protocol, the training set solely consists of drugs drawn from $\mathcal{D}_{\text{I.D.}}$, while validation and test sets include drug pairs where at least one drug belongs to $\mathcal{D}_{\text{O.O.D.}}$.
{
\small
\begin{equation}
\begin{aligned}
\mathcal{D}_{\text{train}} &= \{(d_i, d_j, c_k) \mid d_i, d_j \in \mathcal{D}_{\text{I.D.}},\; c_k \in \mathcal{C}\}, \\
\mathcal{D}_{\text{valid}} \cup \mathcal{D}_{\text{test}} &=
\{(d_i, d_j, c_k) \mid d_i \in \mathcal{D}_{\text{O.O.D.}} \lor d_j \in \mathcal{D}_{\text{O.O.D.}}, c_k \in \mathcal{C}\}.
\end{aligned}
\end{equation}
}

\subsection{Target-Adaptive Disentangled Molecular Graph Encoding}
\label{module1}

To capture both intrinsic molecular topological structures and target-dependent characteristics, we propose the \emph{target-adaptive disentangled molecular graph encoding} model. We first learn target-relevant and target-irrelevant molecular representations through \emph{disentangled molecular graph encoding}, then condition the target-relevant representations on various cell targets through cross-attention with associated target proteins by \emph{target-adaptive representation learning}. To further encourage target-adaptive disentanglement, we explicitly impose a decorrelation constraint on the conditioned target-relevant representations so that they can well preserve different target-specific information. 

\paragraph{Disentangled Molecular Graph Encoding} Each drug molecule $d$ is represented as a molecular graph $\mathcal{G}_d = (\mathcal{V}_d, \mathcal{E}_d)$ where $\mathcal{V}_d$ and $\mathcal{E}_d$ denote the node set and edge set respectively. For each node $v \in \mathcal{V}_d$, we extract a set of node-level representations from $M$ heterogeneous GNNs, and aggregate them into a graph-level vector:
{
\small
\begin{equation}
\mathbf{z}_{\mathcal{G}_d}
=
\Phi_{\text{pool}}
\Big(
\sum_{v \in \mathcal{V}_d}
\Psi\big(
\mathrm{GNN}_1(v), \ldots, \mathrm{GNN}_M(v)
\big)
\Big),
\end{equation}
}
where $\Psi(\cdot)$ denotes a feature fusion operator over multiple GNN views, and $\Phi_{\text{pool}}(\cdot)$ represents the pooling function. The resulting embedding $\mathbf{z}_{\mathcal{G}_d} \in \mathbb{R}^{D}$ captures structural and chemical characteristics of the molecular graph. To separate intrinsic molecular characteristics from target-dependent characteristics, we introduce a disentanglement head that decomposes $\mathbf{z}_{\mathcal{G}_d}$ into target-irrelevant representations and target-relevant representations:
{
\small
\begin{equation}
\mathbf{z}^{\text{irr}}_d = \mathbf{W}_{\text{irr}} \mathbf{z}_{\mathcal{G}_d}, \qquad
\mathbf{z}^{\text{rel}}_d = \mathbf{W}_{\text{rel}} \mathbf{z}_{\mathcal{G}_d},
\end{equation}
}
where $\mathbf{z}^{\text{irr}}_d \in \mathbb{R}^{D_{\text{irr}}}$ captures intrinsic properties of the molecular graph that are independent of specific target protein instantiations, while $\mathbf{z}^{\text{rel}}_d \in \mathbb{R}^{D_{\text{rel}}}$ carries the target-relevant information, with
$D_{\text{irr}} + D_{\text{rel}} = D$.

\paragraph{Target-Adaptive Representation Learning} Target proteins play a critical role in drug synergy prediction, as they mediate the molecular mechanisms through which drugs exert their effects. To incorporate target-specific biological context, we further condition $\mathbf{z}^{\text{rel}}_d$ on drug-associated targets. Let $\{\mathbf{t}^{(k)}\}_{k=1}^{K}$ denote the embeddings of the $K$ corresponding targets related to drug $d$, obtained from a pretrained protein encoder ESM-2~\cite{ESM-2}. We apply a cross-attention mechanism between $\mathbf{z}^{\text{rel}}_d$ and $\{\mathbf{t}^{(k)}\}_{k=1}^{K}$ to produce target-adaptive representations:
{
\small
\begin{equation}
\tilde{\mathbf{z}}^{(k)}_d = \mathrm{CrossAttn}\!\left(\mathbf{z}^{\text{rel}}_d, \mathbf{t}^{(k)}\right), \quad k = 1, \ldots, K.
\end{equation}
}
The final molecular representations are formed by concatenating the target-irrelevant representations with all target-adaptive representations,
{
\small
\begin{equation}
\mathbf{e}_d = \left[ \mathbf{z}^{\text{irr}}_d \,\Vert\, \tilde{\mathbf{z}}^{(1)}_d \,\Vert\, \cdots \,\Vert\, \tilde{\mathbf{z}}^{(K)}_d \right].
\end{equation}
}
To further encourage disentanglement across different target conditioned representations, we impose a decorrelation constraint directly on the set of target-adaptive representations. Specifically, for $K$ target-adaptive representations $\{\tilde{\mathbf{z}}^{(k)}_d\}_{k=1}^{K}$, we explicitly penalize statistical dependencies between every pair of these representations. For $(k, k')$ with $k \neq k'$, we first normalize the representations within a mini-batch and compute their cross-correlation matrix:
{
\small
\begin{equation}
\mathbf{C}_{k,k'} =
\frac{1}{D_k}
\,
\tilde{\mathbf{z}}^{(k)}
\left(\tilde{\mathbf{z}}^{(k')}\right)^{\top},
\end{equation}
}
where $D_k$ denotes the dimensionality of each target-adaptive representation. The decorrelation loss is then defined as the average squared Frobenius norm of the cross-correlation matrices over all pairs:
{
\small
\begin{equation}
\mathcal{L}_{\text{decorr}}
=
\frac{1}{K (K - 1)}
\sum_{k \neq k'}
\left\| \mathbf{C}_{k,k'} \right\|_F^2.
\end{equation}
}
By minimizing $\mathcal{L}_{\text{decorr}}$, the model is encouraged to learn representations that encode distinct aspects of target-relevant drug information, thereby reducing redundancy in representation spaces. This design enforces functional disentanglement across target-irrelevant and target-relevant representations under different cell target contexts, and plays a critical role in improving generalization under drug-level O.O.D. settings.


\subsection{Pairwise Attentive Graph Architecture Search}
\label{module2}

Given new pairs of drugs, we propose the \emph{pairwise attentive graph architecture search} algorithm to discover the optimal GNN architectures for calculating drug molecular graph representations accurately. 
There are three core parts for completing this task: i) \emph{molecular pairwise attention} that injects bidirectional drug pair context into molecular graph representations, ii) \emph{latent operator space parameterization} that projects candidate message-passing operators into a continuous and differentiable latent space, and iii) \emph{adaptive routing for architecture search} that dynamically selects and assembles the most appropriate operators based on molecular graph representations in the projected latent space.

\paragraph{Molecular Pairwise Attention}
Let $\mathbf{e}_{d_1}$ and $\mathbf{e}_{d_2}$ denote molecular graph representations of the two drugs (i.e., $d_1$ and $d_2$) obtained in Sec \ref{module1}. To incorporate the influence of one molecule on the other, we introduce a bidirectional drug pair context injection module based on multi-head attention. Specifically, the representation of drug $d_1$ is expressed as follows:
{
\small
\begin{equation}
\mathbf{h}_{d_1}
= \mathbf{e}_{d_1} +
\mathrm{FFN} ~\big(
\mathcal{A}_{\mathrm{mh}}(\mathbf{Q}, \mathbf{K}, \mathbf{V})
\big),
\end{equation}
}
where the query is constructed by $\mathbf{Q} = \mathbf{W}_Q \mathbf{e}_{d_1}$, while the key-value pairs are derived from $d_2$ as $\mathbf{K}, \mathbf{V} = \mathbf{W}_{K,V} \mathbf{e}_{d_2}$.
An analogous operation is applied symmetrically to obtain $\mathbf{h}_{d_2}$.

Building upon the target-adaptive disentangled molecular graph representations from Sec~\ref{module1}, this attention mechanism allows the pairwise attentive representations to encode pharmacophoric signals that are critical for extrapolating to drug combinations under distribution shifts.

\paragraph{Latent Operator Space Parameterization}
To support pairwise attentive architectural customization while preserving differentiable trainability, we introduce a latent parameterization of candidate message-passing operators. At each GNN layer $l$, we maintain a collection of aggregation primitives $\mathcal{O}^{(l)} = \{\mathrm{op}_1^{(l)}, \mathrm{op}_2^{(l)}, \ldots, \mathrm{op}_m^{(l)}\}$,
where each primitive encodes a distinct strategy for information propagation in molecular graph. We project them into a shared latent operator space by associating each operator $\mathrm{op}_i^{(l)}$ with a learnable vector $\mathbf{o}_i^{(l)} \in \mathbb{R}^{d_{\mathrm{op}}}$. The resulting set $\mathcal{E}^{(l)} = \{\mathbf{o}_1^{(l)}, \ldots, \mathbf{o}_m^{(l)}\}$ defines a continuous operator space that enables smooth interpolation between aggregation patterns.

To ensure that different operators remain functionally distinguishable within this latent space, we explicitly discourage excessive similarity among their representations. Specifically, we introduce a layerwise separation constraint that penalizes high cosine similarity between distinct operator descriptors:
{
\small
\begin{equation}
\mathcal{L}_{\mathrm{sep}}^{(l)} =
\frac{1}{m(m-1)} \sum_{i \neq j}
\left(
\frac{\mathbf{o}_i^{(l)}}{\|\mathbf{o}_i^{(l)}\|_2}
\cdot
\frac{\mathbf{o}_j^{(l)}}{\|\mathbf{o}_j^{(l)}\|_2}
\right)^{2}.
\end{equation}
}
This regularization term softly enforces angular separation among operator representations, preventing the latent operator space from collapsing into a low-rank configuration. The overall $\mathcal{L}_{\mathrm{sep}}$ is obtained by averaging $\mathcal{L}_{\mathrm{sep}}^{(l)}$
across all layers.

\paragraph{Adaptive Routing for Architecture Search}
Given the drug representation $\mathbf{h}_{d}$, we modulate the contribution of each candidate operator at layer $l$ through an adaptive routing mechanism. The routing weight assigned to operator $\mathrm{op}_i^{(l)}$ is computed as follows:
{
\small
\begin{equation}
\alpha_i^{(l)} =
\frac{\exp\left( \langle \mathbf{h}_{d}, \mathbf{o}_i^{(l)} \rangle \right)}
{\sum_{j=1}^{m} \exp\left( \langle \mathbf{h}_{d}, \mathbf{o}_j^{(l)} \rangle \right)},
\end{equation}
}
where $\langle \cdot, \cdot \rangle$ denotes the dot-product similarity. These routing weights can be used to select the best operators for assembling the optimal graph neural architecture. As such, molecular graph representations of the corresponding drugs can be computed through the \textit{optimal} graph neural network as follows:

{
\small
\begin{equation}
\mathbf{x}_{d}^{(l+1)} =
\sum_{i=1}^{m} \alpha_i^{(l)} \,
\mathrm{op}_i^{(l)}\!\left(\mathbf{x}_{d}^{(l)}\right),
\qquad
\mathbf{x}_{d}^{(0)} = \tilde{\mathbf{z}}_v ,
\end{equation}
}
where $\mathbf{x}_{d}^{(l+1)}$ denotes the computed molecule representation at layer $l$. This algorithm allows operator parameters and routing weights to be jointly optimized during finetuning, effectively inducing pair-aware computational graphs.

By conditioning operator routing on pairwise attentive molecular graph representations, this adaptive mechanism prevents the model from over-specializing
to spurious correlations observed in the training distribution.
As a result, the learned representations $\mathbf{x}_{d_1}$ and
$\mathbf{x}_{d_2}$ capture not only intrinsic molecular structures,
but also pair-aware structural adaptations that are capable of
generalization to O.O.D. settings for new drugs.

\subsection{Multi-Level Contextualized Cellular Feature Alignment}
\label{module3}
Drug synergy is inherently cell line dependent, as cellular environments determine drug sensitivity, pathway activation, and synergistic effect. To explicitly incorporate cellular context into drug representations and LLM finetuning, we introduce the \emph{multi-level contextualized cellular feature alignment} mechanism. At the \emph{structural level}, molecular graph representations are augmented with cell line information in the form of context feature concatenation. At the \emph{semantic level}, cellular textual descriptions and gene expression profiles are utilized to align with the LLM input space.

\paragraph{Structural Level Feature Alignment.}
We utilize cell line gene expression profiles to capture context-dependent atomic interactions when learning molecular graph representation. Given a cell line $c$, let $\mathbf{x}_c \in \mathbb{R}^{D_c}$ denote its gene expression feature vector, which is transformed into context representations $\mathbf{e}_c$ via a projection function: $\mathbf{e}_c = \phi_{\text{ctx}}(\mathbf{x}_c)$. For drug $d$ characterized via a molecular graph $\mathcal{G}_d = (\mathcal{V}_d, \mathcal{E}_d)$, each atom in drug $d$ can be described  by node $v \in \mathcal{V}_d$, which is associated with an initial atomic feature vector $\mathbf{z}_v$. We augment the molecular graph representation by concatenating the context representations with atomic feature vectors as follows:
{
\small
\begin{equation}
\tilde{\mathbf{z}}_v = \left[ \mathbf{z}_v \,\Vert\, \mathbf{e}_c \right], \quad \forall v \in \mathcal{V}_d.
\end{equation}
}
This results in a contextualized molecular graph ${\mathcal{G}}_d^{cell}$, in which the atom (i.e., node) representations contain the cellular contexts. Such structural level contextualization allows downstream graph encoders to take the influence of cell specific biological contexts into consideration.

\paragraph{Semantic Level Feature Alignment.}
In addition to structural level concatenation, we further align cellular information at the semantic level by projecting cell line descriptions into the input space of the LLM. Specifically, given textual descriptions of cell line $c$ together with its gene expression vector $\mathbf{x}_c$, we employ a tokenizer and a BERT-based projector~\cite{blip2} to obtain a set of cell specific tokens and representations:
{
\small
\begin{equation}
\mathbf{T}_c = \mathrm{Tokenizer}(c), \qquad
\mathbf{E}_c = \phi_{\text{proj}_c}(\mathbf{x}_c),
\end{equation}
}
where $\mathbf{T}_c$ denotes the discrete cell tokens and $\mathbf{E}_c$ represents continuous cell representations aligned with the input space of the LLM. 

By jointly leveraging structural and semantic level representation contextualization, we are able to align cellular contexts to molecular graphs and language models simultaneously, which is a critical support for O.O.D. generalized DSP for new drugs.

\subsection{DrugSyn-LLM with Retrieval-Augmented Biomedical Instruction Tuning}
\label{module4}

We finetune DrugSyn-LLM, our biomedical LLM, with the proposed \emph{Retrieval-Augmented Biomedical Instruction Tuning} strategy to inject various domain knowledge into LLM while enabling task-specific reasoning. Given an input prompt $\mathcal{P}$, the tokenized textual descriptions of the cell line $\mathbf{T}_c$, its projected representations $\mathbf{E}_c$, the tokenized SMILES sequences of two drugs $\mathbf{T}_{\mathrm{SMILES}_1}$ and $\mathbf{T}_{\mathrm{SMILES}_2}$, DrugSyn-LLM aims to produce an appropriate response $R$ with accurate prediction.

To bridge molecular graph representations and natural language reasoning, the drug representations $\mathbf{x}_{d_1}$ and $\mathbf{x}_{d_2}$ obtained in Sec \ref{module3} are first projected into the language-aligned space via $\phi_{\mathrm{proj}_d}$, which adopts the same BERT-based architecture as $\phi_{\mathrm{proj}_c}$ but is parameterized independently, yielding $\mathbf{E}_{d_1} = \phi_{\mathrm{proj}_d}(\mathbf{x}_{d_1})$ and $\mathbf{E}_{d_2} = \phi_{\mathrm{proj}_d}(\mathbf{x}_{d_2})$.
The overall training procedure consists of two stages: biomedical instruction tuning and task-specific training.

\paragraph{Stage I: Biomedical Instruction Tuning}
In the instruction tuning stage, we aim to ground the LLM with biomedical knowledge relevant to drug and cell contexts. For each training instance, we first retrieve domain-specific biomedical knowledge from curated databases. The retrieved information is organized into a structured biomedical description and utilized by the target response $R$. The instruction prompt $\mathcal{P}_{inst}$ is then deliberately designed to query, explain, or summarize such biomedical evidence, guiding the language model to generate expert level domain knowledge.

During instruction tuning, the large language model is optimized to reproduce the retrieved biomedical descriptions conditioned on the corresponding prompts. Formally, the training objective maximizes the likelihood of generating the target response $R$ in the autoregressive token space as follows:
{
\small
\begin{equation}
\mathcal{L}_{\mathrm{inst}} =
- \log p\!\left( R \mid \mathcal{P}_{inst}, \mathbf{T}_c, \mathbf{T}_{\mathrm{SMILES}_1}, \mathbf{T}_{\mathrm{SMILES}_2} \right),
\end{equation}
}
which encourages the model to internalize pharmacological and biological priors through retrieval-augmented supervision.

\paragraph{Stage II: Task-Specific Training}
In the second stage, we adapt the instruction-tuned DrugSyn-LLM to the DSP task.
The prompt $\mathcal{P}_{task}$ is instantiated with task-oriented instructions, while the expected response $R$ corresponds to the predicted category and its associated synergy score.
In addition to textual inputs, the projected representations $\mathbf{E}_c$, $\mathbf{E}_{d_1}$, and $\mathbf{E}_{d_2}$ are injected into the LLM as auxiliary continuous representations.
The task training objective is defined in the generative output space of the language model.
Specifically, the model is optimized to maximize the likelihood of producing a structured task response $R = \{R_{\mathrm{label}}, R_{\mathrm{score}}\}$, which encodes both the synergy type and the corresponding synergy score as follows:
{
\small
\begin{equation}
\mathcal{L}_{\mathrm{task}} =
- \log p\!\left(
R
\mid
\mathcal{P}_{task},
\mathbf{T}_c,
\mathbf{T}_{\mathrm{SMILES}_1}, 
\mathbf{T}_{\mathrm{SMILES}_2},
\mathbf{E}_c,
\mathbf{E}_{d_1},
\mathbf{E}_{d_2}
\right).
\end{equation}
}

\spara{Putting All Together} The overall training objective combines the generative task loss with representation and architecture regularization terms as follows:
{
\small
\begin{equation}
\label{all}
\mathcal{L} =
\delta \, \mathcal{L}_{\mathrm{inst}}
+ (1 - \delta) \, \mathcal{L}_{\mathrm{task}}
+ \alpha \, \mathcal{L}_{\mathrm{decorr}}
+ \beta \, \mathcal{L}_{\mathrm{sep}},
\end{equation}
}
where $\delta$ is a stage indicator that activates the instruction-tuning objective, i.e., $\delta = 1$ during biomedical instruction tuning and $\delta = 0$ during task-specific training.

\section{Experiment}

We conduct extensive experiments under two O.O.D. settings for both synergy classification and score regression to evaluate the effectiveness of our proposed \model model. Comparisons against a wide range of baselines demonstrate that our method consistently achieves superior performance under distribution shifts.

\renewcommand{\arraystretch}{0.75}
\begin{table}[htbp]
\centering
\caption{O.O.D. Dataset Splitting Statistics based on Scaffold and Size.
$\theta_{\text{scaffold}}$ and $\theta_{\text{size}}$ denote the splitting thresholds
used to partition the dataset into in-distribution ($\mathcal{D}_{\text{I.D.}}$)
and out-of-distribution ($\mathcal{D}_{\text{O.O.D.}}$) subsets.}
\label{tab:dataset}
\resizebox{\linewidth}{!}{%
\small
\begin{tabular}{l|cccccc}
\toprule
\multicolumn{7}{c}{\textbf{Scaffold-based Splitting}} \\
\midrule
Dataset & $\theta_{\text{scaffold}}$\ (mol) & $\mathcal{D}_{\text{I.D.}}$ & $\mathcal{D}_{\text{O.O.D.}}$ &
$\mathcal{D}_{\text{train}}$ & $\mathcal{D}_{\text{valid}}$ & $\mathcal{D}_{\text{test}}$ \\
\midrule
Bliss  & 13 & 1364 & 424 & 84766  & 20391 & 20391 \\
Hsa    & 13 & 912  & 243 & 72830  & 17864 & 17865 \\
Loewe  & 9  & 1775 & 272 & 109676 & 27070 & 27070 \\
Zip    & 13 & 1320 & 413 & 59513  & 14767 & 14767 \\
\midrule
\multicolumn{7}{c}{\textbf{Size-based Splitting}} \\
\midrule
Dataset & $\theta_{\text{size}}$\ (Da) & $\mathcal{D}_{\text{I.D.}}$ & $\mathcal{D}_{\text{O.O.D.}}$ &
$\mathcal{D}_{\text{train}}$ & $\mathcal{D}_{\text{valid}}$ & $\mathcal{D}_{\text{test}}$ \\
\midrule
Bliss  & 305 & 1305 & 480 & 83318  & 21115 & 21115 \\
Hsa    & 305 & 878  & 277 & 71964  & 18302 & 18302 \\
Loewe  & 260 & 1753 & 294 & 112276 & 25770 & 25770 \\
Zip    & 300 & 1295 & 438 & 60330  & 14358 & 14359 \\
\bottomrule
\end{tabular}
}
\end{table}
\vspace{-4pt}

\subsection{Settings}

\begin{figure}[htbp]
  \centering
  \includegraphics[width=1\linewidth]{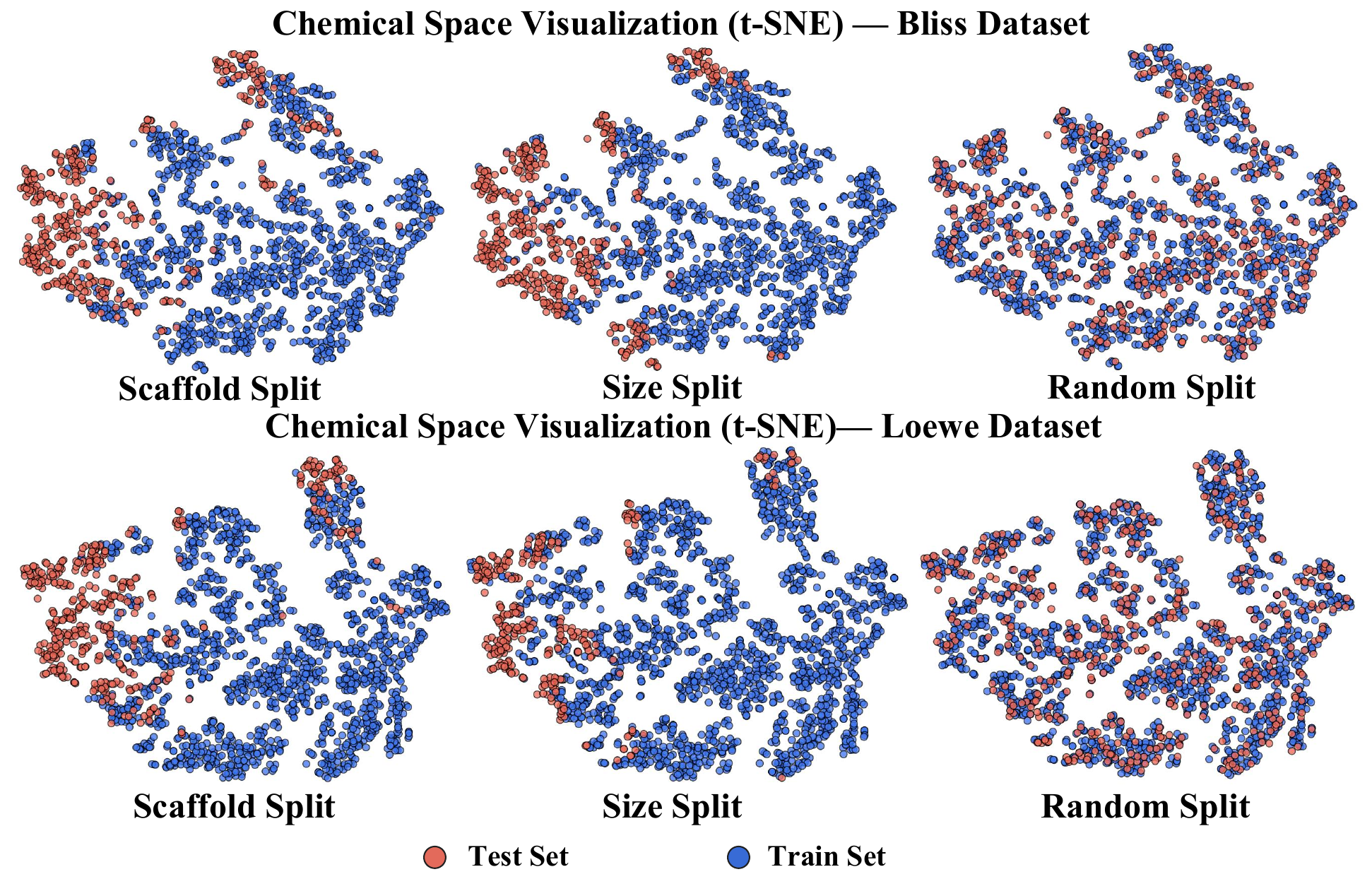}
  \caption{Chemical space visualization.}
  \label{fig:pca}
\end{figure}

\begin{table*}[h]
\setlength\tabcolsep{1.5pt} 
\renewcommand{\arraystretch}{0.75} 
\caption{Comparative performance on scaffold-based and size-based O.O.D. DSP classification and regression tasks. The top-performing method is highlighted in bold. We also report the percentage improvement of these metrics compared to the second-best performing method. ``-'' indicates that the corresponding method does not support this task.}
\label{tab:performance}
\begin{center}
\resizebox{\linewidth}{!}{%
\small
\begin{tabular}{c l cccc cccc cccc cccc}
\toprule
Setting & Model 
& \multicolumn{4}{c}{Bliss}
& \multicolumn{4}{c}{HSA}
& \multicolumn{4}{c}{Loewe} 
& \multicolumn{4}{c}{ZIP} \\
\cmidrule(l{\tabcolsep}){3-6} \cmidrule(l{\tabcolsep}){7-10} \cmidrule(l{\tabcolsep}){11-14} \cmidrule(l{\tabcolsep}){15-18}
&& ACC $\uparrow$ & AUC $\uparrow$ & MAE $\downarrow$ & RMSE $\downarrow$
& ACC $\uparrow$ & AUC $\uparrow$ & MAE $\downarrow$ & RMSE $\downarrow$
& ACC $\uparrow$ & AUC $\uparrow$ & MAE $\downarrow$ & RMSE $\downarrow$
& ACC $\uparrow$ & AUC $\uparrow$ & MAE $\downarrow$ & RMSE $\downarrow$ \\
\midrule
\multicolumn{18}{c}{\textbf{Scaffold-Based O.O.D. Drug Synergy Prediction Tasks}} \\
\midrule
\multirow{9}{*}{\makecell{DNN-\\Based}}  
& DeepSynergy & 64.53 & 72.14 & 24.83 & 34.05 & 72.88 & 75.62 & 14.19 & 18.61 & 93.01 & 82.28 & 13.31 & 17.73 & 67.16 & 75.61 & 17.73 & 22.32 \\
& DFFNDDS & 55.98 & 47.75 & - & - & 66.04 & 49.58 & - & - & 80.74 & 53.59 & - & - & 49.91 & 48.17 & - & - \\
& TranSynergy & 54.95 & 61.21 & 26.69 & 34.41 & 76.79 & 73.14 & 14.51 & 18.74 & 92.93 & 77.66 & 12.62 & 17.51 & 57.46 & 65.16 & 17.42 & 21.17 \\
& MatchMaker & 59.80 & 62.92 & 25.56 & 34.29 & 77.97 & 73.71 & 14.28 & 18.52 & 93.11 & 78.55 & 12.35 & 17.33 & 57.49 & 67.35 & 17.07 & 20.71 \\
& TreeCombo & 61.80 & 68.70 & 28.08 & 43.62 & 72.35 & 76.32 & 15.70 & 19.51 & 93.05 & 80.14 & 14.52 & 22.90 & 65.65 & 75.23 & 16.24 & 19.93 \\
& MarSY & 58.51 & 61.32 & 27.17 & 36.07 & 74.67 & 73.18 & 15.42 & 19.39 & 92.92 & 77.22 & 12.36 & 17.71 & 55.19 & 66.21 & 17.35 & 20.89 \\
& DTSyn & 56.89 & 59.79 & - & - & 74.42 & 69.49 & - & - & 63.08 & 77.67 & - & - & 62.03 & 69.22 & - & - \\
& SynergyX & - & - & 27.76 & 35.28 & - & - & 15.58 & 19.34 & - & - & 13.82 & 19.60 & - & - & 17.51 & 21.16 \\
\cmidrule(l{\tabcolsep}){1-18}
\multirow{6}{*}{\makecell{GNN-\\Based}}   
& DeepDDS & 62.10 & 66.82 & - & - & 68.70 & 73.95 & - & - & 91.18 & 74.30 & - & - & 66.61 & 74.21 & - & - \\
& DDoS & 63.71 & 70.26 & - & - & 71.16 & 73.03 & - & - & 91.53 & 76.38 & - & - & 68.97 & 76.27 & - & - \\
& GAECDS & 61.54 & 61.82 & - & - & 71.23 & 72.56 & - & - & 91.93 & 78.94 & - & - & 65.96 & 71.90 & - & - \\
& JointSyn & 59.07 & 62.86 & 26.56 & 34.41 & 77.26 & 69.62 & 14.73 & 18.99 & 92.81 & 73.56 & 12.59 & 18.01 & 62.36 & 71.09 & 16.16 & 20.33 \\
& MFSynDCP & 65.01 & 70.60 & - & - & 67.31 & 73.42 & - & - & 75.21 & 70.93 & - & - & 68.83 & 76.07 & - & - \\
& AttenSyn & 55.91 & 58.51 & - & - & 76.62 & 70.90 & - & - & 92.76 & 72.73 & - & - & 58.00 & 61.15 & - & - \\
\cmidrule(l{\tabcolsep}){1-18}
\multirow{3}{*}{\makecell{LLM-\\Based}}  
& CancerGPT & 71.74 & 81.17 & 21.74 & 32.63 & 77.65 & 80.66 & 13.74 & 18.72 & 89.52 & 85.64 & 13.49 & 19.30 & 76.08 & 78.11 & 14.21 & 19.29 \\
& BAITSAO & 68.29 & 75.18 & 24.77 & 33.28 & 75.13 & 79.85 & 15.07 & 18.88 & 91.61 & 80.42 & 14.01 & 19.18 & 68.03 & 76.21 & 15.02 & 19.00 \\
\cmidrule(l{\tabcolsep}){3-18}
& \textbf{\modelnosp} 
& \textbf{77.27} & \textbf{85.31} & \textbf{20.63} & \textbf{29.24}
& \textbf{80.98} & \textbf{83.48} & \textbf{11.74} & \textbf{16.91}
& \textbf{93.55} & \textbf{86.19} & \textbf{10.09} & \textbf{15.80}
& \textbf{76.98} & \textbf{85.84} & \textbf{11.56} & \textbf{17.17} \\
& \%$\uparrow$ & \textbf{+7.71\%} & \textbf{+5.10\%} & \textbf{-5.11\%} & \textbf{-10.39\%}
& \textbf{+3.86\%} & \textbf{+3.50\%} & \textbf{-14.56\%} & \textbf{-8.69\%}
& \textbf{+0.54\%} & \textbf{+0.64\%} & \textbf{-25.20\%} & \textbf{-12.27\%}
& \textbf{+1.18\%} & \textbf{+9.90\%} & \textbf{-18.65\%} & \textbf{-9.63\%} \\

\midrule
\multicolumn{18}{c}{\textbf{Size-Based O.O.D. Drug Synergy Prediction Tasks}} \\
\midrule
\multirow{9}{*}{\makecell{DNN-\\Based}}  
& DeepSynergy & 68.31 & 78.05 & 27.46 & 35.86 & 75.85 & 75.95 & 14.74 & 18.57 & 91.31 & 78.07 & 12.99 & 17.67 & 73.34 & 83.12 & 17.04 & 24.07 \\
& DFFNDDS & 54.51 & 50.95 & - & - & 68.40 & 50.78 & - & - & 87.46 & 57.84 & - & - & 51.32 & 50.67 & - & - \\
& TranSynergy & 59.31 & 62.45 & 26.66 & 35.82 & 76.48 & 77.46 & 14.53 & 18.43 & 92.86 & 76.96 & 12.19 & 17.03 & 61.65 & 65.17 & 17.92 & 22.39 \\
& MatchMaker & 58.18 & 62.66 & 26.87 & 35.58 & 75.76 & 71.80 & 14.39 & 19.21 & 92.72 & 76.54 & 12.13 & 17.06 & 63.98 & 69.97 & 17.30 & 22.02 \\
& TreeCombo & 62.54 & 67.87 & 27.74 & 39.09 & 69.38 & 73.39 & 15.83 & 19.97 & 92.60 & 78.35 & 13.58 & 19.43 & 68.22 & 76.03 & 16.22 & 20.53 \\
& MarSY & 56.73 & 59.26 & 27.97 & 36.85  & 74.11 & 70.35 & 14.91 & 19.53 & 92.86 & 77.82 & 12.13 & 17.33 & 53.98 & 68.49 & 18.10 & 22.21 \\
& DTSyn & 54.76 & 57.33 & - & - & 73.01 & 68.03 & - & - & 92.46 & 77.66 & - & - & 64.28 & 69.40 & - & - \\
& SynergyX & - & - & 26.95 & 35.98 & - & - & 14.69 & 19.50 & - & - & 12.48 & 18.08 & - & - & 18.35 & 22.56 \\
\cmidrule(l{\tabcolsep}){1-18}
\multirow{6}{*}{\makecell{GNN-\\Based}}   
& DeepDDS & 65.69 & 71.29 & - & - & 71.92 & 77.07 & - & - & 87.98 & 79.25 & - & - & 68.06 & 74.09 & - & - \\
& DDoS & 66.90 & 73.37 & - & - & 72.00 & 75.37 & - & - & 90.48 & 72.49 & - & - & 69.06 & 74.09 & - & - \\
& GAECDS & 58.54 & 62.28 & - & - & 71.65 & 70.23 & - & - & 90.48 & 79.09 & - & - & 60.87 & 72.00 & - & - \\
& JointSyn & 59.12 & 62.40 & 26.60 & 34.45 & 76.08 & 66.98 & 15.27 & 19.78 & 92.70 & 79.61 & 12.23 & 17.74 & 63.03 & 70.14 & 16.62 & 21.35 \\
& MFSynDCP & 65.28 & 69.85 & - & - & 65.99 & 70.61 & - & - & 77.10 & 71.35 & - & - & 65.94 & 71.89 & - & - \\
& AttenSyn & 53.86 & 56.38 & - & - & 72.47 & 67.60 & - & - & 92.76 & 78.71 & - & - & 61.93 & 65.36 & - & - \\
\cmidrule(l{\tabcolsep}){1-18}
\multirow{3}{*}{\makecell{LLM-\\Based}}  
& CancerGPT & 71.92 & 80.30 & 23.97 & 35.39 & 77.75 & 78.14 & 14.00 & 19.10 & 93.17 & 85.23 & 11.66 & 17.47 & 73.99 & 81.75 & 14.07 & 21.65 \\
& BAITSAO & 69.38 & 76.91 & 24.36 & 33.13 & 76.59 & 79.63 & 14.96 & 18.79 & 92.65 & 80.67 & 12.65 & 17.57 & 69.39 & 76.40 & 15.93 & 21.19 \\
\cmidrule(l{\tabcolsep}){3-18}
& \textbf{\modelnosp}
& \textbf{77.66} & \textbf{85.79} & \textbf{20.85} & \textbf{30.05}
& \textbf{79.97} & \textbf{83.00} & \textbf{10.95} & \textbf{16.30}
& \textbf{96.17} & \textbf{96.80} & \textbf{10.42} & \textbf{16.00}
& \textbf{76.56} & \textbf{85.25} & \textbf{12.66} & \textbf{19.72} \\
& \%$\uparrow$ & \textbf{+7.98\%} & \textbf{+6.84\%} & \textbf{-13.02\%} & \textbf{-9.30\%}
& \textbf{+2.86\%} & \textbf{+4.23\%} & \textbf{-21.79\%} & \textbf{-11.56\%}
& \textbf{+3.22\%} & \textbf{+13.58\%} & \textbf{-10.63\%} & \textbf{-6.21\%}
& \textbf{+3.47\%} & \textbf{+2.56\%} & \textbf{-10.02\%} & \textbf{-3.95\%} \\

\bottomrule
\end{tabular}}
\end{center}
\end{table*}

\paragraph{Dataset} 
We derive all drug combination data from DrugComb~\cite{drugcomb}, which contains totally 1,432,351 unique <drug, drug, cell line> triplets. Each triplet is annotated with synergy measurements under four scoring schemes, namely Loewe, Bliss, HSA, and ZIP. Detailed definitions and computation rules for these synergy scores are provided in Appendix A. Drug-related information and features is primarily obtained from DrugBank~\cite{drugbank}, while gene expression profiles of cell lines are collected from the CancerRx-Gene~\cite{cancerrx} database. 
We first filter the samples following the recommendation of the SynergyFinder software~\cite{synergyfinder} documentation, retaining only drug combinations that exhibit pronounced synergistic or antagonistic effects ($\lvert \text{score} \rvert \geq 10$). Subsequently, the filtered dataset is partitioned into $\mathcal{D}_{\text{I.D.}}$ and $\mathcal{D}_{\text{O.O.D.}}$ according to the protocol described in Sec ~\ref{module0}, where \emph{scaffold} refers to the core chemical framework of a molecule that defines its structural backbone, while \emph{size} denotes the molecular weight of the compound. Specifically, domains with descriptor values exceeding a predefined threshold $\theta$ are assigned to the training split, while the remaining domains are used for validation and testing, following common practice~\cite{drugood}. Training, validation, and test sets are organized with an approximate ratio of $4{:}1{:}1$. Dtailed statistics are reported in Table~\ref{tab:dataset}.

\paragraph{Distribution Analysis.}
Based on a set of generic chemical space descriptors, we visualize the molecular distributions of the Bliss and Loewe datasets under different splitting strategies using t-SNE.
As illustrated in Fig.~\ref{fig:pca}, the proposed O.O.D. split yields a clear separation between the train set ($\mathcal{D}_{\mathrm{I.D.}}$) and the test set ($\mathcal{D}_{\mathrm{O.O.D.}}$) across multiple chemical dimensions.
This separation substantially increases the difficulty of model generalization, as test compounds reside in chemically distinct regions from those observed during training.
In contrast, conventional random splitting, although introducing unseen drugs in the test set, leads to strong overlap and coupling between seen and unseen compounds in chemical space, thereby failing to provide a reliable assessment of a model’s generalization capability. 

\paragraph{Baselines}
We perform extensive benchmarking against three categories of baseline methods: (i) Conventional DNN-based models: DeepSynergy~\cite{DeepSynergy}, DFFNDDS~\cite{DFFNDDS}, TranSynergy~\cite{TranSynergy}, MatchMaker~\cite{MatchMaker}, TreeCombo~\cite{TreeCombo}, MarSY~\cite{MarSY}, DTSyn~\cite{DTSyn}, and SynergyX~\cite{SynergyX}; (ii) GNN-based methods: DeepDDS~\cite{DeepDDS}, DDoS~\cite{DDoS}, GAECDS~\cite{GAECDS}, JointSyn~\cite{JointSyn}, MFSynDCP~\cite{MFSynDCP}, and AttenSyn~\cite{AttenSyn}; (iii) State-of-the-art LLM-based approaches: CancerGPT~\cite{CancerGPT} and BAITSAO~\cite{BAITSAO}. The comparisons across multiple paradigms ensure rigorous evaluations under diverse architectural and learning settings.

\paragraph{Drug Descriptors}

We retrieve drug-related information from the DrugBank~\cite{drugbank} database, including SMILES sequence and basic physicochemical properties for each drug. These drug descriptors are aligned with the downstream drug–drug–cell line triplets using drug names as a common identifier. The SMILES sequences are further processed using the RDKit~\cite{rdkit} toolkit to construct graph-structured molecular representations, where atoms and bonds are modeled as nodes and edges. In addition to molecular structures, other retrieved drug attributes are incorporated through carefully designed prompts and corresponding target outputs, which are used during the instruction tuning stage in Sec \ref{module4} to enhance the model’s ability to leverage heterogeneous drug knowledge.

\paragraph{Cell Line Representations}

We obtain cell line features from the CancerRx-Gene~\cite{cancerrx} resource, which provides Robust Multi-array Average (RMA)–normalized~\cite{rma} basal gene expression profiles for approximately 1000 human cancer cell lines. Each cell line is originally characterized by genome-wide transcriptional measurements covering 17,737 genes. In this study, we focus on a subset of 908 landmark genes curated by the L1000 project~\cite{L1000}.

\paragraph{Protein Representations.}
We obtain the amino acid sequences of target proteins from UniProt~\cite{uniprot} and encode them using ESM-2~\cite{ESM-2}. 
ESM-2 is a large-scale protein language model pre-trained on millions of protein sequences, which captures rich evolutionary and structural information through self-supervised learning.

\paragraph{Implementation Details}
For classification settings, we evaluate model performance using Accuracy and Area Under the Receiver Operating Characteristic Curve (AUC-ROC). For regression tasks, we adopt Mean Absolute Error (MAE) and Root Mean Squared Error (RMSE) as evaluation metrics. We employ galactica~\cite{Galactica} pre-trained on large-scale scientific corpora as the backbone LLM architecture.  In addition, the cross-modal projection layers are initialized with representations derived from SciBERT~\cite{scibert}, providing a semantically informed starting point for multimodal alignment. Most experiments are conducted on NVIDIA A100-SXM4 GPUs with 40\,GB memory.

\subsection{Results}

As shown in Table~\ref{tab:performance}, \model consistently outperforms all other baselines across all metrics, datasets, and settings. This demonstrates the superior performance of \model in addressing O.O.D. generalized DSP. We have the following key observations:

\spara{Classification Results} i) The classification metrics on Loewe are consistently higher than those on others. This phenomenon can be attributed to the severe class imbalance in Loewe, under which conditions AUC provides a more reliable measure of discriminative performance. Notably, our method achieves a clear advantage on this metric, highlighting its superior classification capability despite the biased label distribution. ii) LLM-based methods demonstrate superior advantages, suggesting that large language models can leverage semantic information to generalize to O.O.D. drug pairs. By explicitly retrieving and injecting domain-specific medical knowledge, our method achieves a clear performance margin over generic LLM-based approaches. iii) DNN-based and GNN-based methods exhibit no substantial difference in overall performance. Notably, some even underperform DeepSynergy~\cite{DeepSynergy}, which relies on drug fingerprints, in O.O.D. settings. This suggests that incorporating excessive or highly complex features may introduce significant noise and redundancy, leading models to capture spurious correlations that fail to generalize beyond the training distributions.

\spara{Regression Results} i) Regression constitutes a more challenging task, as it requires accurate modeling of fine-grained numerical outcomes rather than coarse decision boundaries. Compared with corresponding baselines, our method exhibits a markedly larger performance margin, providing stronger evidence of its effectiveness and accuracy. ii) Several methods are not inherently designed to handle both tasks in a unified manner or are limited to a single prediction paradigm, while others derive classification outcomes indirectly from regression value ranges, which may introduce evaluation bias and compromise result fidelity. Our method is capable of simultaneously generating both outputs in a chain-of-thought manner, enabling coherent reasoning across tasks, which endows \model with greater flexibility and scalability.

\begin{figure}[htbp]
  \centering
  \includegraphics[width=1\linewidth]{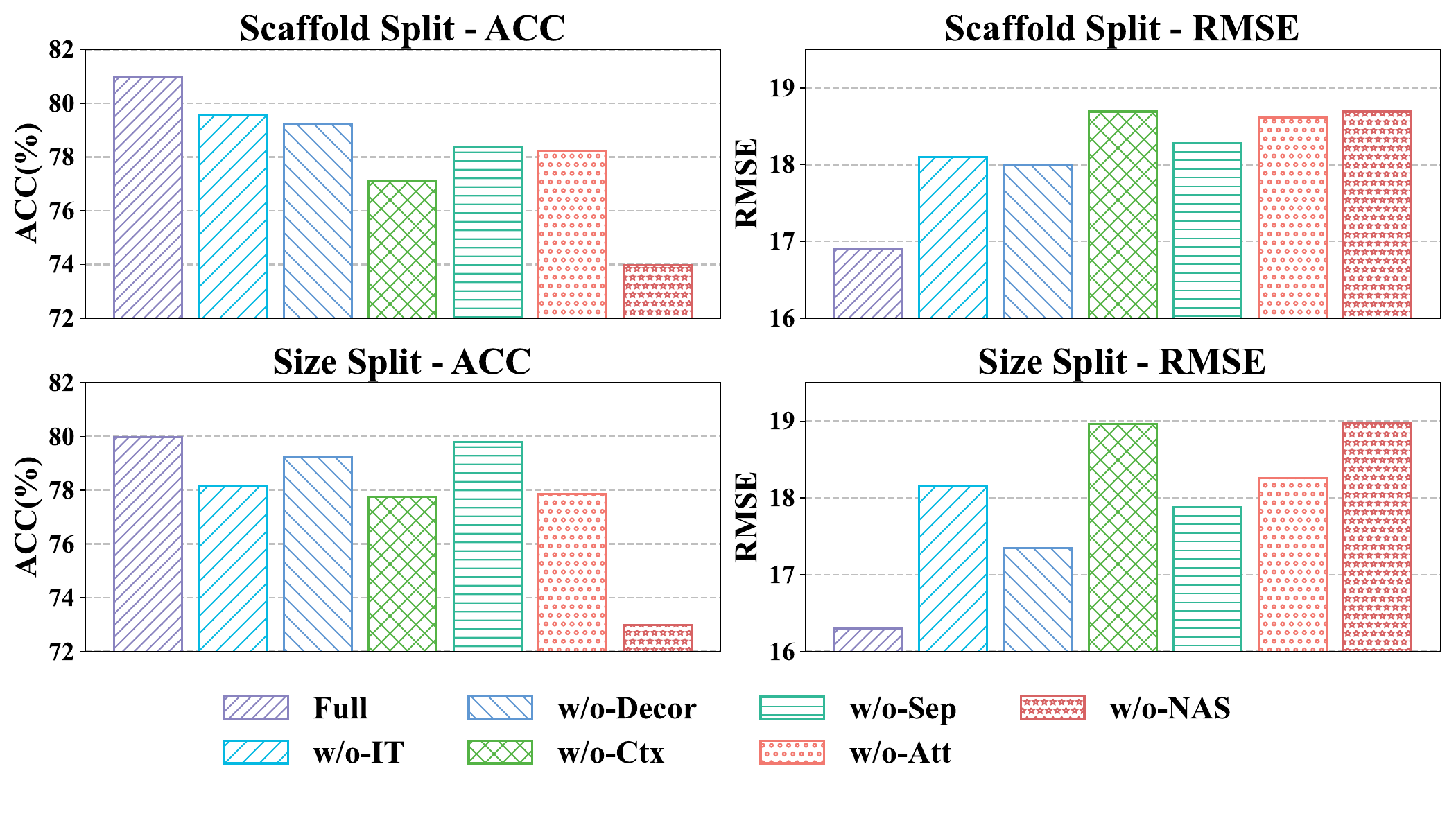}
  \vspace{-8mm}
  \caption{Ablation studies of \modelnosp.}
  \label{fig:ablation}
\end{figure}
\vspace{-2mm}

\subsection{Ablation Study}

To assess the impact of individual components within our \modelnosp, we perform an ablation study on HSA score dataset under both splits. Model performance is evaluated using accuracy for classification and RMSE for regression. We design multiple model variants, where each variant excludes a particular component:

\begin{itemize}[leftmargin=0.5cm]
    \item \textbf{w/o R-IT}: We remove the retrieval-augmented biomedical instruction tuning and directly perform task-specific training without external knowledge retrieval. %

    \item \textbf{w/o Decor}: We discard the decorrelation constraint $\mathcal{L}_{\text{decorr}}$, thus removing the enforcement of disentanglement among target-conditioned drug representations.

    \item \textbf{w/o NAS}: We disable the neural architecture search process and instead adopt a handcrafted graph neural network to represent molecular structures. %

    \item \textbf{w/o Attn}: We eliminate the pairwise attention mechanism and guide the architecture search using independent drug representations solely. 

    \item \textbf{w/o Sep}: We remove the separation constraint $\mathcal{L}_{\text{sep}}$, which enforces the dispersion of operation representations.

    \item \textbf{w/o Ctx}: We exclude structural level cell line features as the contexts and only utilize raw molecular graph topological features.
\end{itemize}

The ablation results are summarized in Fig.~\ref{fig:ablation}. 
Removing any single component consistently leads to noticeable performance degradation across different metrics, indicating that each component contributes meaningfully to our \modelnosp. 
Among all variants, \textit{w/o Ctx} and \textit{w/o NAS} incur the most pronounced performance drops, particularly on the more challenging regression task. 
This observation highlights the importance of explicitly modeling cell line contextual information, as well as tailoring neural architectures based on target-adaptive molecular encodings.

\subsection{Hyperparameter Analysis}

\begin{figure}[htbp]
  \centering
  \includegraphics[width=1\linewidth]{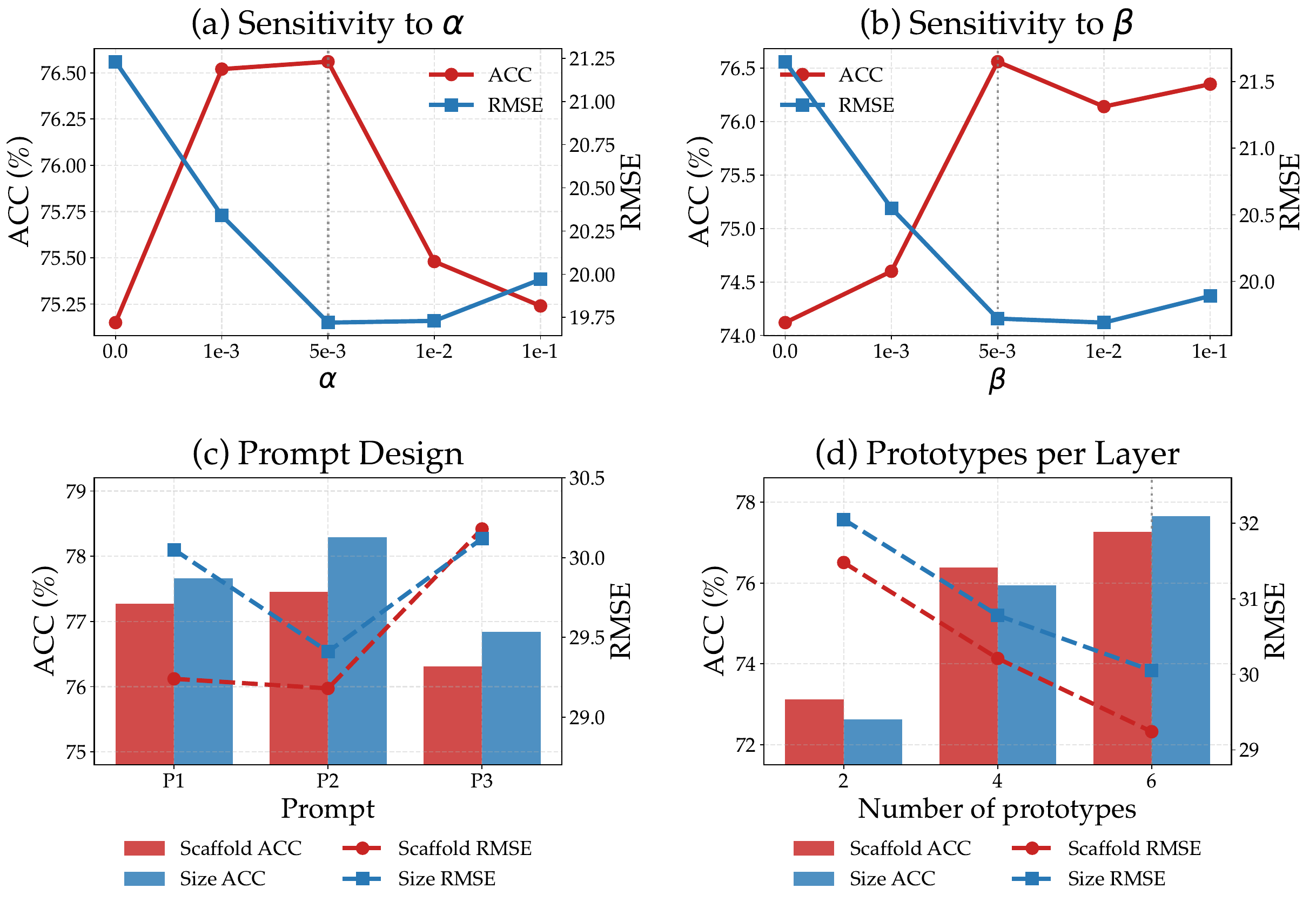}
  \caption{Hyperparameter sensitivity analysis for $\alpha$, $\beta$, different prompts and prototypes per layer.}
  \label{fig:hyper}
\end{figure}

To investigate the sensitivity of our framework to key hyperparameters, we analyze the impact of varying $\alpha$ and $\beta$ in Eq.~\ref{all} on both classification and regression performance in the size-based splitted zip dataset. Specifically, we consider $\alpha, \beta \in \{0.0, 1\mathrm{e}{-3}, 5\mathrm{e}{-3}, 1\mathrm{e}{-2}, 1\mathrm{e}{-1}\}$, and report the results in Fig.~\ref{fig:hyper}. 

As shown in the figure, the model achieves the best overall performance when $\alpha = 5\mathrm{e}{-3}$, while increasing or decreasing $\alpha$ leads to noticeable performance degradation across different tasks. In contrast, increasing $\beta$ beyond this range has a relatively minor effect on both accuracy and RMSE, whereas reducing $\beta$ results in a more pronounced decline. Consistent with the ablation study, removing either loss term corresponding to $\alpha$ or $\beta$ causes a significant drop in performance, highlighting the necessity of both components in our objective.

We further evaluate the robustness of our framework to different prompt formulations. 
We consider three prompts under the same experimental setting on the Bliss dataset:
\textbf{P1}: ``Do the two drugs exhibit synergy effects? What is their [score] synergy score?'';
\textbf{P2}: ``Classify the synergy effects between the two drugs and report their [score] synergy score.'';
and \textbf{P3}: ``As a pharmacovigilance officer, how would you classify and calculate the [score] synergy score between the two drugs?''.
The performance remains stable across these prompt variants, suggesting that our framework is not overly sensitive to surface-level prompt wording. 
This robustness indicates that the model primarily relies on the learned structural, contextual, and semantic representations rather than exploiting a specific prompt template.

Finally, we analyze the effect of the number of prototypes per layer in the architecture search module. 
This hyperparameter controls the capacity of the architecture search space by determining how many prototype operations are maintained at each layer. 
The results show that increasing the number of prototypes from $2$ to $6$ consistently improves both classification and regression performance. 
We use $6$ prototypes per layer as the default setting, which achieves the best overall performance in our experiments.

\subsection{Case Study}

\begin{figure}[htbp]
  \centering
  \includegraphics[width=0.9\linewidth]{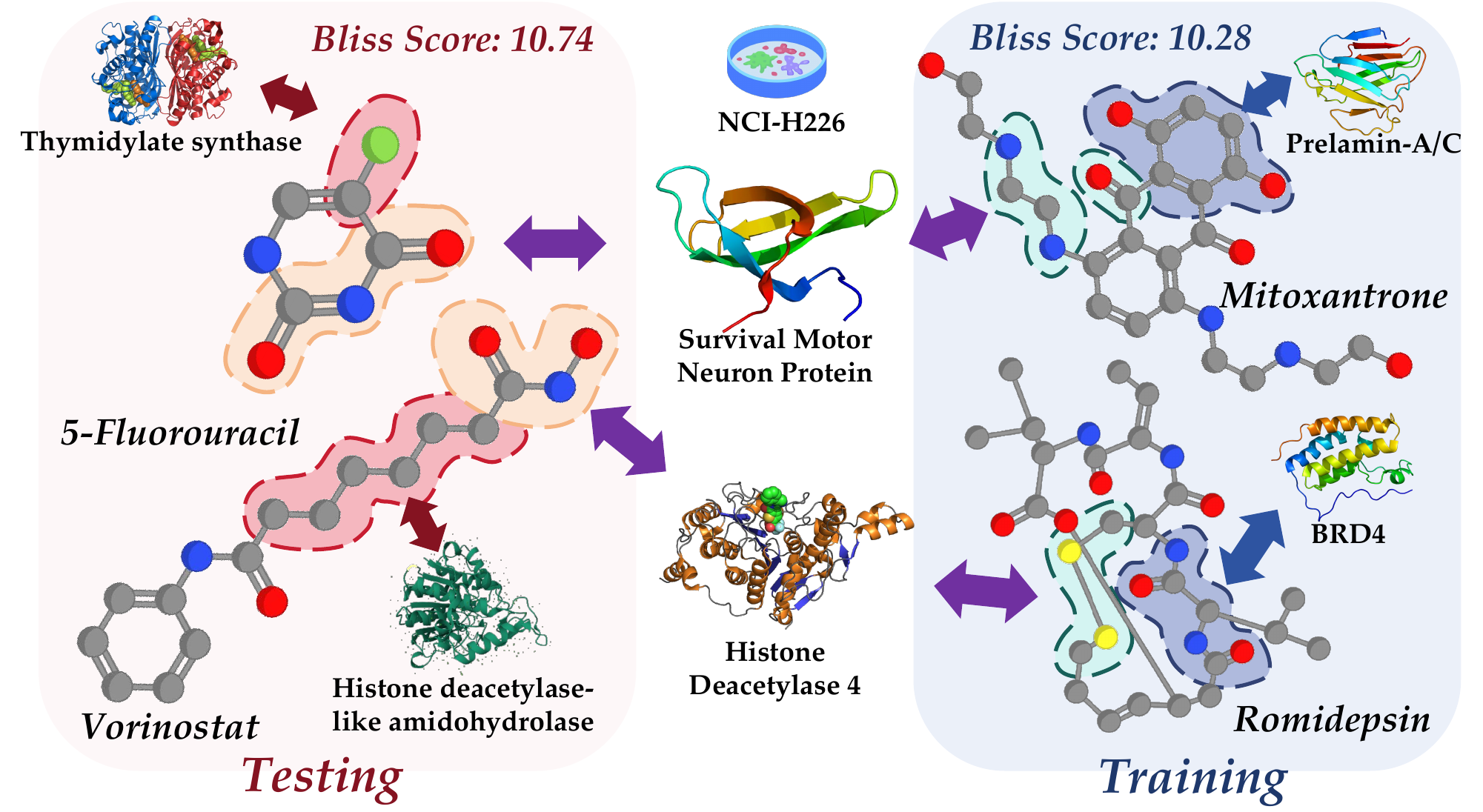}
  \caption{A case study on 5-Fluorouracil and Vorinostat.}
  \label{fig:study}
\end{figure}

We further present a case study on 5-Fluorouracil and Vorinostat, whose molecular scaffolds and sizes both exhibit substantial topological shifts relative to the training distributions, in the NCI-H226 cell line under Bliss score. 

Specifically, as shown in Fig.~\ref{fig:study}, 5-Fluorouracil and Mitoxantrone share a target-conditioned clue related to the survival motor neuron (SMN) protein. 
Although structurally distinct, both contain polar carbonyl-bearing planar motifs that can support hydrogen-bond-mediated interactions around SMN-associated protein-RNA interfaces, suggesting a shared biochemical basis for modulating SMN complex stability rather than acting through direct enzymatic inhibition.
This shared polar interaction pattern provides a transferable SMN-related signal for O.O.D. inference. 
Similarly, Vorinostat and Romidepsin converge on histone deacetylase regulation, particularly HDAC4 and related zinc-dependent HDAC isoforms. 
Vorinostat inhibits HDACs through hydroxamate-based \ce{Zn^{2+}} chelation, whereas Romidepsin, after intracellular reduction, exposes a thiol group that serves an analogous metal-coordinating role. 
These shared SMN- and HDAC4-centered clues indicate that our target-adaptive modeling can align mechanistically related signals across structurally divergent drugs, thereby supporting generalization under O.O.D. settings.

\section{Conclusion}
In this work, we propose \modelnosp, a novel graph LLM framework for out-of-distribution (O.O.D.) generalized drug synergy prediction (DSP) that unifies molecular graph representation and biomedical semantic language representations through a joint optimization. 
Extensive experiments 
demonstrate that the proposed OOD-GraphLLM consistently outperforms state-of-the-art approaches on various DSP tasks.
To the best of our knowledge, \model is the first attempt to study O.O.D. generalized DSP by resorting to graph large language models.

\section*{Acknowledgement}
This work was supported by the National Key Research and Development Program of China No.2023YFF1205001.

\bibliographystyle{ACM-Reference-Format}
\bibliography{ref}

\appendix

\section{Experiment Details}

\subsection{Operations}
\label{appendix:operations}

To enable flexible architecture search over molecular graph encoders, we define a candidate operator set $\mathcal{O}^{(l)}$ at each layer $l$. 
All operators are implemented as bond-aware message-passing functions, where node features $x_i$ and bond features $e_{ij}$ are jointly used to construct molecular messages.

\begin{itemize}
    \item \textbf{\texttt{GCNmol}.} 
    A GCN-style operator that combines neighboring node features with bond features and performs degree-normalized aggregation.

    \item \textbf{\texttt{GINmol}.} 
    A GIN-style operator that applies an MLP to the central node feature and aggregated bond-aware neighbor messages:
    {\small
    \begin{equation}
        h_i = \mathrm{MLP}\left((1+\epsilon)x_i 
        + \sum_{j \in \mathcal{N}(i)} \mathrm{ReLU}(x_j + e_{ij})\right).
    \end{equation}
    }

    \item \textbf{\texttt{GATmol}.} 
    An attention-based operator that weights bond-aware neighbor messages with attention coefficients $\alpha_{ij}$:
    {\small
    \begin{equation}
        h_i = \sum_{j \in \mathcal{N}(i)} 
        \alpha_{ij}\,\mathrm{ReLU}(x_j + e_{ij}).
    \end{equation}
    }

    \item \textbf{\texttt{SAGEmol}.} 
    A GraphSAGE-style operator that aggregates bond-aware neighborhood information and fuses it with a transformed root-node representation.

    \item \textbf{\texttt{Graphmol}.} 
    A GraphConv-style operator that applies separate transformations to the central node and aggregated neighbor messages before combining them.

    \item \textbf{\texttt{MLPmol}.} 
    A graph-agnostic operator that only applies a learnable linear transformation to the current node feature, serving as a skip-like transformation candidate.
\end{itemize}

All candidate operators share a unified interface. 
During architecture search, the model learns a weighted combination over these operators, allowing each layer to adaptively select suitable message-passing functions for molecular representation learning.

\subsection{Prompt Design}
\label{appendix:prompt}

\begin{figure}[htbp]
  \centering
  \includegraphics[width=1\linewidth]{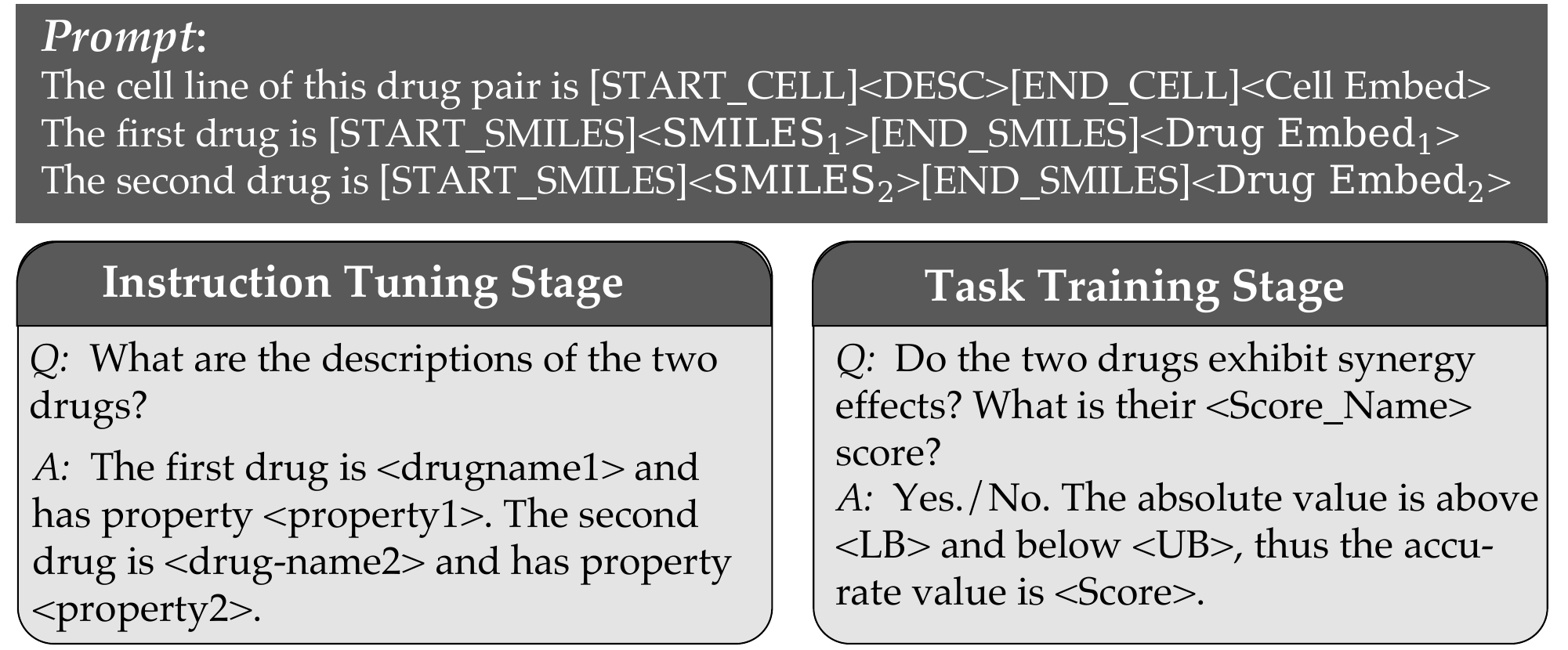}
  \caption{Detailed input prompt design for DrugSyn-LLM.}
  \label{fig:prompt}
\end{figure}
\vspace{-6pt}

As illustrated in Fig.~\ref{fig:prompt}, we carefully design the prompts used to fine-tune DrugSyn-LLM. Specifically, cell line information and drug smiles are incorporated as the backbone prompt to provide sufficient biological and chemical context. During the instruction tuning stage, the question-answer format is centered on describing and reasoning about intrinsic drug properties. In the subsequent task-specific training stage, we adopt a \emph{Chain-of-Thought} (CoT) prompting strategy. The model is first guided to predict the categorical synergy outcome, after which it is prompted to infer a bounded interval by specifying the lower and upper bounds of the synergy score. Finally, the model is instructed to output an exact numerical value within this range.

\subsection{Instruction Tuning Stategy}
\label{appendix:instruction_tuning}

To inject reliable biomedical knowledge into the instruction tuning process, we design the retrieval procedure as a curated and deterministic grounding mechanism, which aims to ensure that the language model receives high-quality and consistent textual knowledge during training.

Specifically, we construct a local drug-description database from DrugBank~\cite{drugbank}. For each drug, we collect its metadata from the \textit{IDENTIFICATION} section, where the \textit{Summary} field is used as the primary textual description. For a small number of drugs with missing or incomplete structured entries in DrugBank, we manually verify and supplement their descriptions using trusted biomedical sources such as ChEMBL~\cite{chembl}.

During instruction tuning, retrieval is performed by exact matching with DrugBank identifiers. Given a drug in a training instance, its DrugBank ID is used to deterministically map the drug to its corresponding description in the local database. This description is directly formatted into the target instruction text as structured biomedical knowledge.

\subsection{Implementation Details}

We train the model using the AdamW~\cite{adam} optimizer with a numerical stability constant of $\epsilon = 1\mathrm{e}{-8}$ and apply weight decay ($\lambda = 0.05$) as a regularization mechanism. The learning rate is governed by a two-stage schedule, where it is first gradually increased from $1\mathrm{e}{-6}$ to $1\mathrm{e}{-4}$ during the warm-up phase, and subsequently reduced following a cosine decay strategy until $1\mathrm{e}{-5}$. Although different parameter groups are assigned distinct base learning rates, they all share the same global scheduling policy. To enable efficient fine-tuning of the LLM, we adopt Low-Rank Adaptation (LoRA)~\cite{lora} with rank $r = 16$, while keeping approximately $99.8\%$ of the original model parameters frozen.

\section{More Analyses}

\subsection{Message Passing}
\label{appendix:message_passing}

To further understand how information propagates during architecture search, we analyze the operation weights and relate them to molecular structural properties. For each molecule occurrence, we extract the learned operation weights and compute their associations with multiple structural properties.

For the correlation between operation weights and structural metrics, we noticed that the operation weights are not uniformly distributed across molecular structures. In particular, \textbf{\texttt{GATmol}} shows a positive Spearman correlation with heavy atom count ($\rho = 0.4701$). Similarly, \textbf{\texttt{GRAPHmol}} is positively correlated with aromatic ring count ($\rho = 0.4495$), suggesting that graph-level aggregation becomes more important for ring-rich molecular structures. 

\begin{figure}[htbp]
    \centering
    \includegraphics[width=0.95\linewidth]{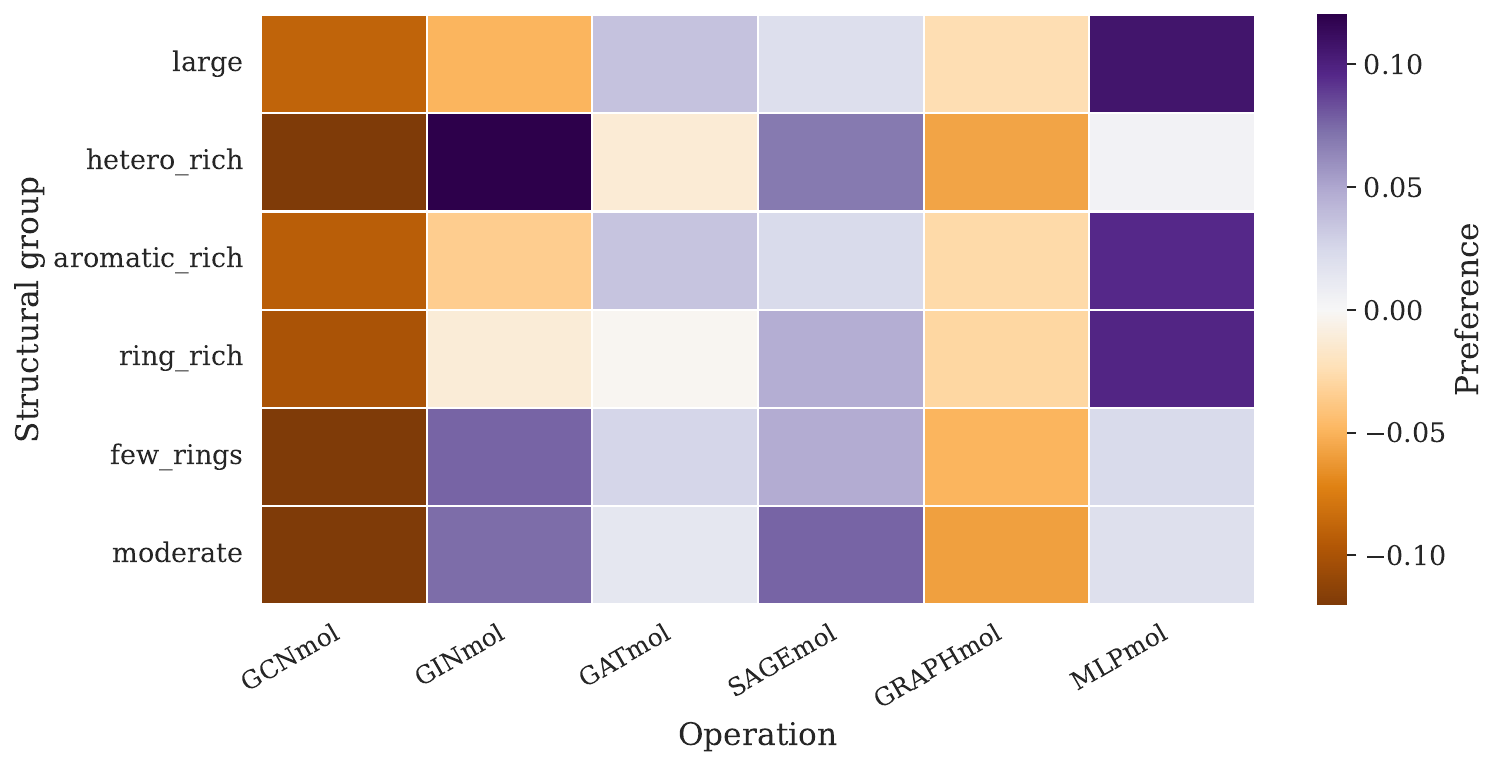}
    \caption{
    Structural-group preferences of diverse message-passing operations.
    Colors indicate relative within-group operation preference after subtracting the mean operation weight of each structural group. Positive values indicate operations emphasized within a group, while negative values indicate relatively suppressed operations.
    }
    \label{fig:operations}
\end{figure}

We visualize operation preferences across molecular structural groups in Fig.~\ref{fig:operations}. The visualization reveals clear structural specialization. Small and hetero-atom-rich molecules show stronger preference for \textbf{\texttt{GINmol}}. In contrast, \textbf{\texttt{GINmol}} is suppressed in ring-rich molecules. Ring-rich molecules instead show increased preference for \textbf{\texttt{MLPmol}}, suggesting that the model relies more on feature transformation and less on \textbf{\texttt{GINmol}}-style local expressive aggregation for ring-rich structures.

Representative molecule-level examples further support this trend. A small hetero-atom-rich molecule, \smiles{CN1C(=O)N2C=NC(=C2N=N1)C(=O)N}, assigns high weight to \textsc{GINmol}, indicating that expressive local aggregation is useful for compact structures with dense hetero-atom patterns. By contrast, a larger aromatic multi-ring molecule, \smiles{CS(=O)(=O)C1=CC(=C(C=C1)C(=O)NC2=CC(=C(C=C2)Cl)C3=CC=CC=N3)Cl}, shows increased weights for \textsc{GATmol} and \textsc{GRAPHmol}, consistent with the need for attention-based and graph-level propagation over larger aromatic systems.


\subsection{Interpretability}
\label{appendix:interpretability}

We further examine whether the internal representations of our framework provide biologically and chemically meaningful signals. We conduct interpretability analyses at two levels: target-level attention and substructure-level SMILES attention.

\paragraph{Target-level interpretability.}
We first inspect the cross-attention scores between drug representations and protein-target embeddings. For each analyzed drug-combination sample, we identify the target receiving the highest attention weight and compare it with known pharmacological mechanisms. Representative examples are summarized in Table~\ref{tab:target_interpretability}. The attended targets are consistent with established biological knowledge in several cases. For example, KU-55933 attends strongly to ATM, which agrees with its known role as an ATM inhibitor involved in DNA damage response. Similarly, Imiquimod attends to Toll-like receptor 7, consistent with its function as a TLR7 agonist. 

\begin{table}[htbp]
\setlength\tabcolsep{1.5pt} 
\renewcommand{\arraystretch}{0.75} 
\centering
\caption{Target-level interpretability examples.}
\label{tab:target_interpretability}
\small
\begin{tabularx}{\linewidth}{YYY}
\toprule
\textbf{Drug Pair} & \textbf{Top Attended Target} & \textbf{Mechanistic Relevance} \\
\midrule
Perifosine + KU-55933 
& Serine-protein kinase ATM 
& ATM inhibition / DNA damage response \\

5-Fluorouracil + Imiquimod 
& Toll-like receptor 7 
& Immune activation via TLR7 signaling \\

Melphalan hydrochloride + Carmustine 
& Glutathione reductase 
& Oxidative stress and detoxification pathways \\
\bottomrule
\end{tabularx}
\end{table}

\paragraph{Substructure-level interpretability.}

\begin{figure}[htbp]
\centering
\includegraphics[width=\linewidth]{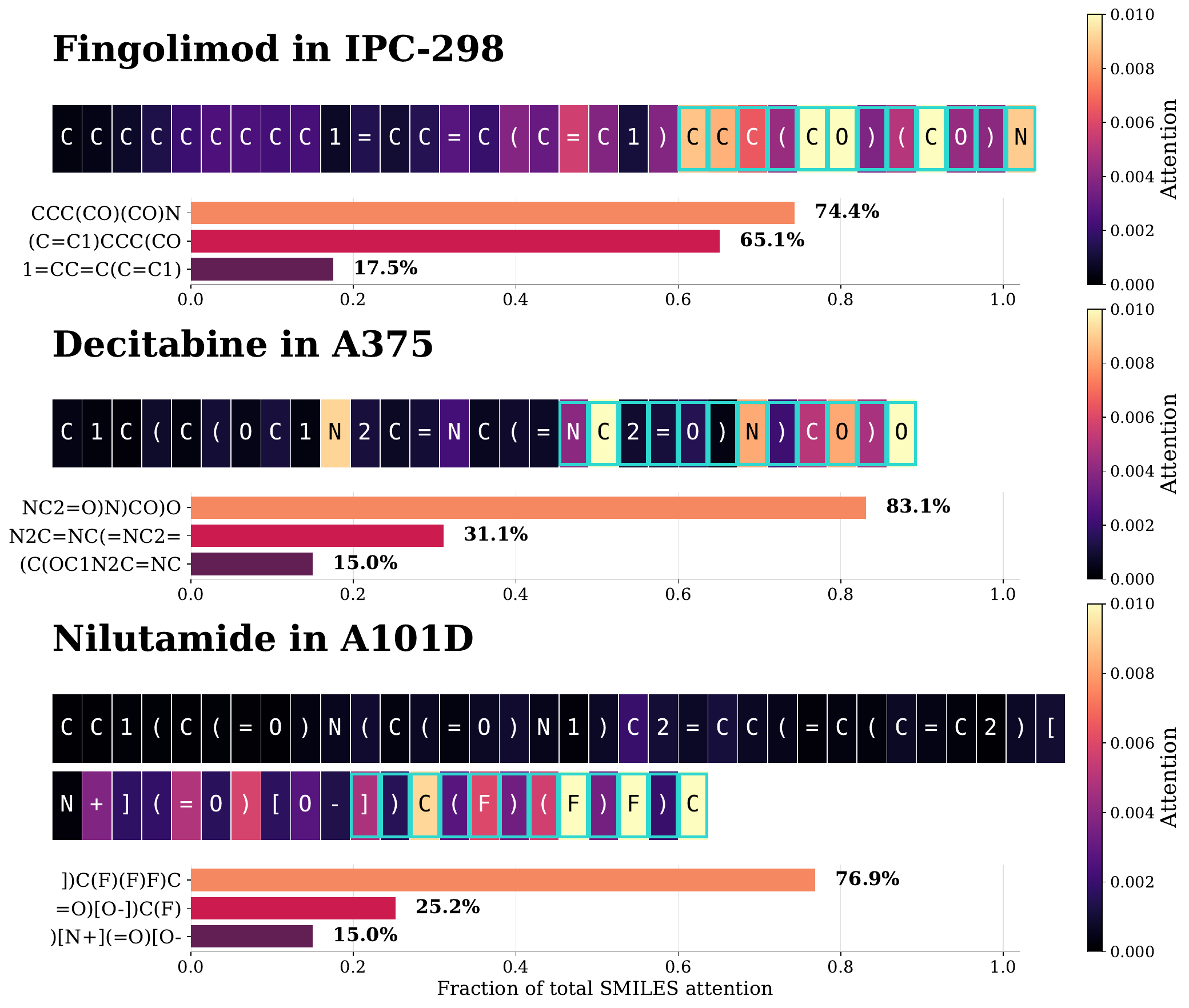}
\caption{
Substructure-level SMILES attention visualization for Fingolimod, Decitabine, and Nilutamide. 
Warmer colors indicate higher character-level attention. Highlighted regions correspond to high-attention local fragments, and the bar plots report their fractions of total SMILES attention.
}
\label{fig:smiles_attention}
\end{figure}

We further analyze whether the model focuses on chemically meaningful molecular substructures. Specifically, we project the final-layer language-model attention scores back to the SMILES characters and identify high-attention local fragments. Figure~\ref{fig:smiles_attention} visualizes three representative drugs. Each panel shows the character-level SMILES attention heatmap, with cyan boxes marking the highest-attention fragment.

The highlighted fragments are chemically interpretable. For Fingolimod, the model assigns high attention to the amino-alcohol-containing region, which is consistent with the polar head-group characteristics of this molecule. For Decitabine, the attention concentrates on an N/O-rich nucleoside-like region, matching its chemically distinctive heteroatom-rich structure. For Nilutamide, the model emphasizes the fluorinated aromatic-side fragment, including the trifluoromethyl-related region, which is relevant to lipophilicity and molecular recognition in medicinal chemistry.

Overall, these results provide supporting evidence that the learned representations capture signals aligned with known pharmacological and medicinal-chemistry knowledge.

\end{document}